\documentclass[letterpaper, 10 pt, conference]{ieeeconf}
\IEEEoverridecommandlockouts
\usepackage{times}
\usepackage{graphicx}
\usepackage{tabularx}
\usepackage{array}
\usepackage{amsmath,amssymb,mathtools}
\usepackage{cite}
\usepackage{balance}
\usepackage[hidelinks]{hyperref}
\usepackage{booktabs}

\newcolumntype{L}[1]{>{\raggedright\arraybackslash}p{#1}}

\newtheorem{assumption}{Assumption}
\newtheorem{property}{Property}

\newtheorem{remark}{Remark}

\title{\LARGE \bf
SEP-NMPC: Safety Enhanced Passivity-Based Nonlinear Model Predictive Control for a UAV Slung Payload System
}

\author{%
Seyedreza Rezaei, Junjie Kang, Amaldev Haridevan, and Jinjun Shan%
\thanks{ This work was supported by the Discovery Grant from the Natural Sciences and Engineering Research Council of Canada (NSERC).}%
\thanks{All authors are with the Department of Earth and Space Science and Engineering, York University, Toronto, ON M3J~1P3, Canada (email: {srrezaei, kjj, amaldev, jjshan}@yorku.ca.) (Corresponding author: Jinjun Shan.)}%
}

\begin{document}
\maketitle
\thispagestyle{empty}
\pagestyle{empty}

\begin{abstract}

Model Predictive Control (MPC) is widely adopted for agile multirotor vehicles, yet achieving both stability and obstacle-free flight is particularly challenging when a payload is suspended beneath the airframe.
This paper introduces a Safety Enhanced Passivity-Based Nonlinear MPC (SEP-NMPC) that provides formal guarantees of stability and safety for a quadrotor transporting a slung payload through cluttered environments.
Stability is enforced by embedding a strict passivity inequality, which is derived from a shaped energy storage function with adaptive damping, directly into the NMPC. This formulation dissipates excess energy and ensures asymptotic convergence despite payload swings.
Safety is guaranteed through high-order control barrier functions (HOCBFs) that render user-defined clearance sets forward-invariant, obliging both the quadrotor and the swinging payload to maintain separation while interacting with static and dynamic obstacles.
The optimization remains quadratic-program compatible and is solved online at each sampling time without gain scheduling or heuristic switching.
Extensive simulations and real-world experiments confirm stable payload transport, collision-free trajectories, and real-time feasibility across all tested scenarios.
The SEP-NMPC framework therefore unifies passivity-based closed-loop stability with HOCBF-based safety guarantees for UAV slung-payload transportation.
Video: \href{https://youtu.be/l04DesGVjwc}{https://youtu.be/l04DesGVjwc}.
\end{abstract}

\section{Introduction}
The past decade has witnessed a rapid rise in the use of unmanned aerial vehicles (UAVs), driven by their flexibility, high maneuverability, and relative ease of deployment. UAVs are now routinely employed in scenarios that are dangerous or inaccessible to humans, including inspection, surveillance, search and rescue, and logistics missions~\cite{8435987,6290694,6299175}. In addition to these roles, UAVs provide an attractive platform for aerial transport, offering the ability to quickly deliver supplies and equipment across environments where ground access is difficult or time-consuming~\cite{9462539}.  

A widely studied approach for aerial transport is the use of cable-suspended, or slung payloads~\cite{Lee2013}. These systems are appealing because they allow a UAV to carry loads without requiring specialized grippers or rigid mechanical attachments. At the same time, the suspended payload introduces significant dynamic complexity: the payload motion is coupled with the airframe dynamics, and the flexible cable adds extra degrees of freedom that render the overall system underactuated. As a result, rapid maneuvers can induce oscillations in the payload, which in turn reduce tracking accuracy and increase the risk of collisions with nearby obstacles, particularly in cluttered or constrained settings~\cite{Tang2015}.  

\begin{figure}[t]
    \centering
    \includegraphics[width=0.9\linewidth]{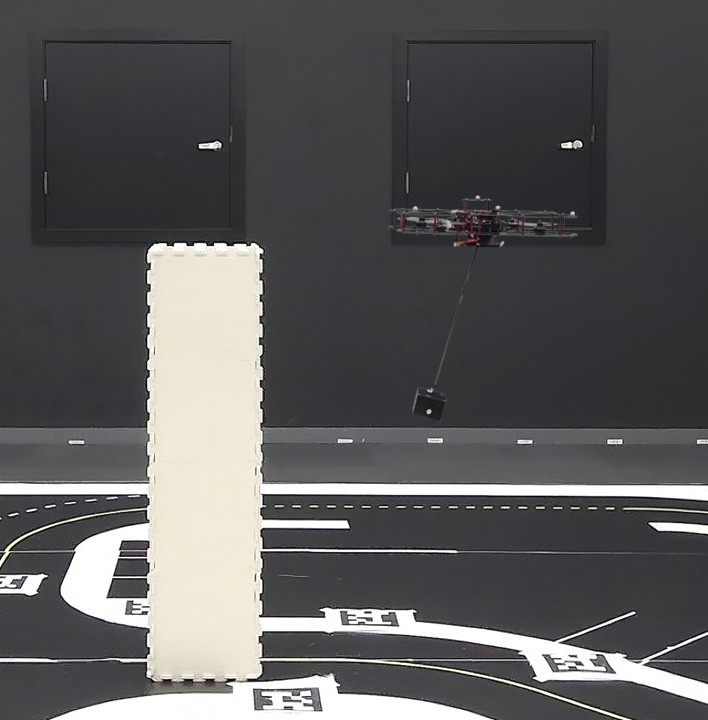}
    \caption{Illustration of a quadrotor with a slung payload navigating in a cluttered environment. 
    The swinging load expands the effective geometry, motivating the need for stability and safety-aware control.}
    \label{fig:intro}
\vspace{-7pt}
\end{figure}

Model Predictive Control (MPC) is widely adopted for UAV slung payload systems as it enables online trajectory optimization with input and state constraints~\cite{Tartaglione2017,Sun2022}. 
Most existing MPC schemes lack closed-loop stability guarantees and ignore the payload’s swinging footprint, complicating safety enforcement and obstacle avoidance~\cite{Foehn2017RSS,9516971,9456981}. 
Pallar et al.~\cite{Pallar2025} address safety through discrete-time CBF-based trajectory planning but provide no closed-loop stability guarantee.
ES-HPC-MPC~\cite{ESHPC2025} enforces exponential stability and handles slack-to-taut cable transitions under perception constraints but does not consider obstacle avoidance or payload clearance in cluttered environments. 
Other NMPC-based approaches for obstacle avoidance typically rely on hard or soft state constraints to keep the system within safe regions, but they do not provide guaranteed forward invariance or formal safety certificates~\cite{tasooji2025eventtriggerednonlinearmodelpredictive,10341785}.

To this end, we propose a Safety Enhanced Passivity-Based NMPC (SEP-NMPC). The method incorporates a shaped energy function that couples position error with payload swing energy, and enforces a strict passivity inequality within the MPC optimization. This dissipates excess energy and ensures asymptotic stability without resorting to gain tuning. In parallel, we design high-order control barrier functions that explicitly account for the relative degree two clearance constraints of both the quadrotor and the suspended payload. By embedding these conditions as affine inequalities within the MPC optimization, the framework guarantees forward invariance of safety sets while remaining computationally tractable for real-time implementation on lightweight onboard processors.

\textbf{Contributions.}  
(i) A passivity-based storage function embedded in NMPC that guarantees asymptotic stability of the underactuated quadrotor–slung payload system, capturing the coupled dynamics between the vehicle and the swinging load.
(ii) A unified safety envelope that guarantees minimum distance separation for the combined UAV slung payload geometry under static and moving obstacles, ensuring clearance for both the vehicle and the swinging load.  
(iii) A QP-compatible SEP-NMPC scheme that integrates stability and safety constraints seamlessly, without heuristic switching, and runs in real time. The framework provides certified performance in challenging scenarios, such as narrow passages and environments with static and dynamic obstacles.

\begin{table}[t]
\vspace{+10pt}
\centering
\caption{Notation (core symbols)}
\vspace{-8pt}
\label{tab:notation}
\setlength{\tabcolsep}{3pt}
\renewcommand{\arraystretch}{1.05}
\scriptsize

\begin{tabularx}{\columnwidth}{@{}l X@{}}
\toprule
\toprule
\textbf{Symbol} & \textbf{Meaning} \\
\midrule
$\mathcal F_I,\ \mathcal F_B$ & Inertial and body-fixed frames \\
$\xi=[x,y,z]^\top,\ \dot\xi$ & Position and velocity in $\mathcal F_I$ \\
$R\in SO(3),\ \omega$ & Attitude; body angular velocity \\
$\gamma=[\alpha,\beta]^\top$ & Payload swing angles ($xz/yz$ planes) \\
$p_Q=\xi,\ p_L$ & Quadrotor / payload positions \\
$m_Q,\ m_L,\ l,\ g$ & Quadrotor/payload masses; cable length; gravity \\
$u_a,\ v=\dot\xi$ & Shaped translational input; collocated output (passivity) \\
$h_{i,j}$ & Squared-clearance CBF (\eqref{eq:cbf_clearance}) \\
$A_{\mathrm{CBF}},\ b_{\mathrm{CBF}}$ & Stacked HOCBF matrices (\eqref{eq:cbf_stack}) \\
$\rho,\ \varepsilon$ & Passivity gains (\eqref{eq:sep_passivity}) \\
$\mathcal U,\ \mathcal A$ & Actuator/swing constraint sets \\
$Q,\ R,\ T,\ N$ & Weights; prediction horizon; shooting nodes \\
\bottomrule
\bottomrule
\end{tabularx}
\vspace{-12pt}
\end{table}

\section{System Dynamics}

\begin{figure}[!t]
\vspace{+8pt}

    \centering
    \includegraphics[width=0.435\linewidth]{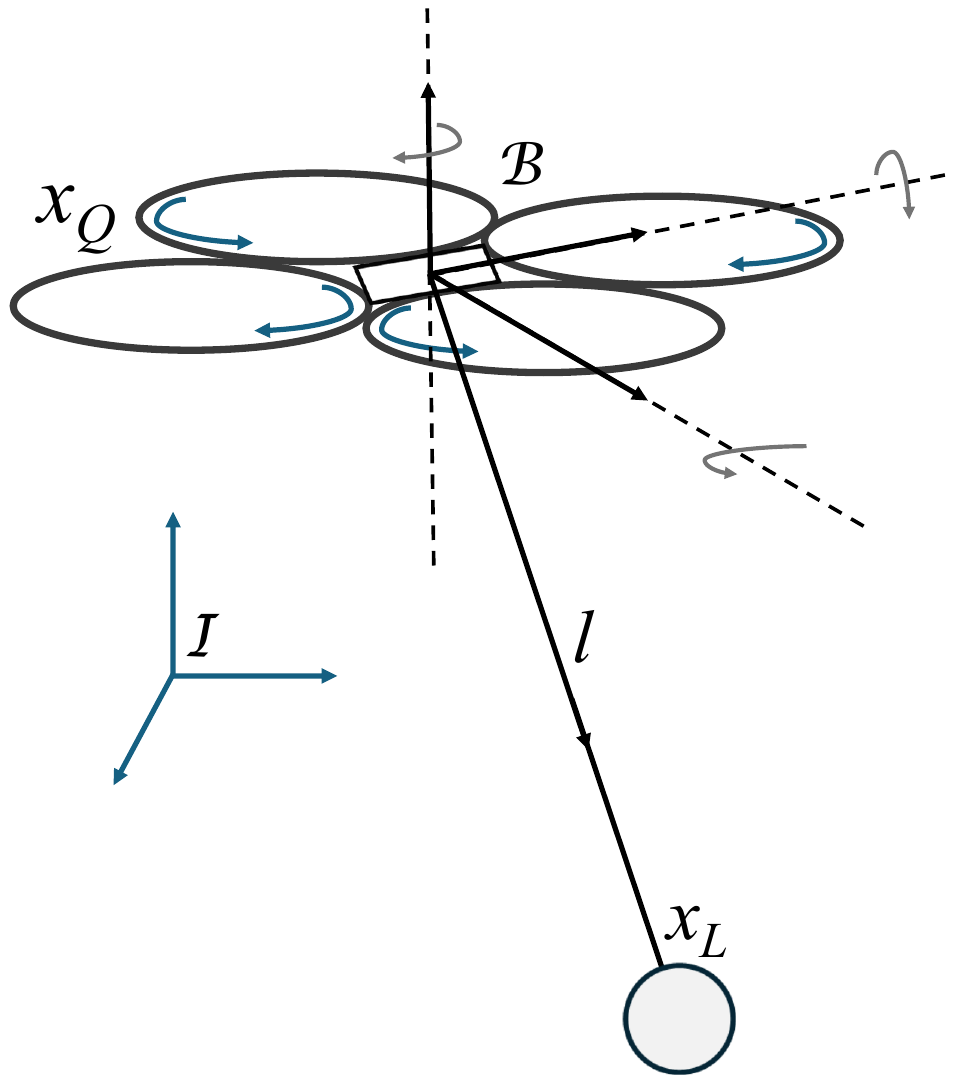}
    \hspace{0.005\linewidth} 
    \includegraphics[width=0.385\linewidth]{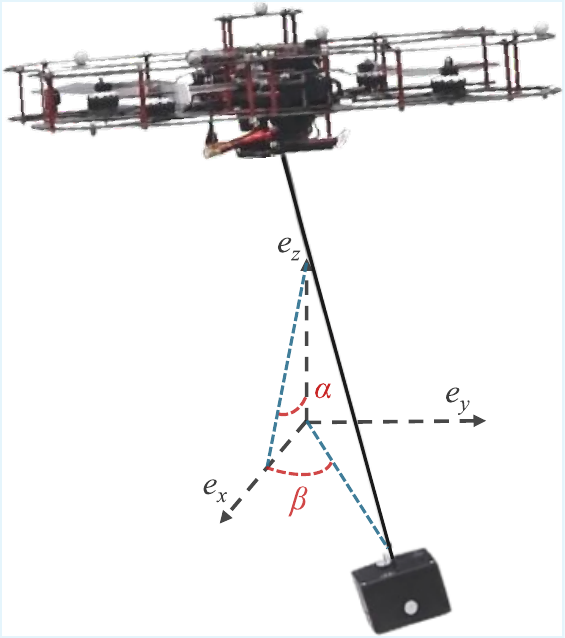}
    \caption{Coordinate frames and configuration of a quadrotor with a slung payload: 
    the inertial frame $\mathcal{F}_I$ with origin $O$, the body-fixed frame $\mathcal{F}_B$, 
    cable length $l$, and swing angles $\alpha$ (in $xz$ plane) and $\beta$ (in $yz$ plane).}
    \label{fig:system_config}
\vspace{-8pt}
\end{figure}

Consider an unmanned quadrotor transporting a point-mass payload suspended by a massless rigid cable of length $l$, as illustrated in Fig.~\ref{fig:system_config}. We define two coordinate frames: the inertial frame $\mathcal{F}_I$ with origin $O$ and the body-fixed frame $\mathcal{F}_B$ attached to the quadrotor's center of mass. The configuration of the system is described by the quadrotor position $\xi = [x, y, z]^T \in \mathbb{R}^3$ in $\mathcal{F}_I$, the quadrotor attitude represented by the rotation matrix $R \in SO(3)$ from $\mathcal{F}_B$ to $\mathcal{F}_I$, and the cable swing angles $\gamma = [\alpha, \beta]^T \in \mathbb{R}^2$, where $\alpha$ and $\beta$ denote the projected swing angles in the $xz$ and $yz$ planes, respectively~\cite{8038248, 10149091}.
The dynamics of the quadrotor-payload transportation system can be derived using the Lagrangian formulation. The system exhibits a cascaded underactuated structure with eight degrees of freedom (DOFs) and only four control inputs. We separate the dynamics into translational (outer-loop) and rotational (inner-loop) subsystems.

\subsubsection{Translational Dynamics}

Let $q = [\xi^\top, \gamma^\top]^\top \in \mathbb{R}^5$ denote the generalized position vector. The translational dynamics can be expressed as:
\begin{equation}
M(q)\ddot{q} + C(q, \dot{q})\dot{q} + G(q) = u
\label{eq:trans_dynamics}
\end{equation}
where $M(q) \in \mathbb{R}^{5 \times 5}$ is the symmetric positive-definite inertia matrix, $C(q, \dot{q}) \in \mathbb{R}^{5 \times 5}$ represents the Coriolis and centrifugal matrix, $G(q) \in \mathbb{R}^5$ is the gravitational force vector, and $u=\operatorname{col}\left(F_x, F_y, F_z, 0,0\right)$, and $F:=\operatorname{col}\left(F_x, F_y, F_z\right) \in \mathbb{R}^3$ denotes the input force actuating the UAV's positions.

The inertia matrix $M(q)$ is given by:
\begin{equation*}
M = \begin{bmatrix}
(m_Q + m_L)I_3 & M_{c} \\
M_{c}^\top & M_{p}
\end{bmatrix}
\end{equation*}
where $m_Q$ and $m_L$ denote the masses of the quadrotor and payload, respectively, $I_3$ is the $3 \times 3$ identity matrix, and:
\begin{align*}
M_{c} &= m_L l\begin{bmatrix}
\cos\alpha\cos\beta & -\sin\alpha\sin\beta \\
0 & \cos\beta \\
\sin\alpha\cos\beta & \cos\alpha\sin\beta
\end{bmatrix} \\
M_{p} &= m_L l^2\begin{bmatrix}
\cos^2\beta & 0 \\
0 & 1
\end{bmatrix}
\end{align*}

The Coriolis and centrifugal matrix is defined as $C = [c_{ij}]_{5 \times 5}$, where $c_{ij} = -c_{ji}$, 
and all elements not listed below are zero:
\begin{equation*}
\begin{aligned}
& c_{14} = -m_L l\, \dot{\alpha} \sin\alpha \cos\beta - m_L l\, \dot{\beta} \cos\alpha \sin\beta, \\
& c_{15} = -m_L l\, \dot{\alpha} \cos\alpha \sin\beta - m_L l\, \dot{\beta} \sin\alpha \cos\beta, \\
& c_{25} = -m_L l\, \dot{\beta} \sin\beta, \\
& c_{34} = m_L l\, \dot{\alpha} \cos\alpha \cos\beta - m_L l\, \dot{\beta} \sin\alpha \sin\beta, \\
& c_{35} = -m_L l\, \dot{\alpha} \sin\alpha \sin\beta + m_L l\, \dot{\beta} \cos\alpha \cos\beta, \\
& c_{44} = -m_L l^2\, \dot{\beta} \cos\beta \sin\beta, \quad
  c_{45} = -m_L l^2\, \dot{\alpha} \cos\beta \sin\beta.
\end{aligned}
\end{equation*}

The gravitational vector, where $g$ denotes the gravitational acceleration, is
\begin{equation}
\begin{aligned}
G = \big[&0,\ 0,\ (m_Q{+}m_L)g,\\
& m_L g l \sin\alpha \cos\beta,\ m_L g l \cos\alpha \sin\beta \big]^\top .
\end{aligned}
\end{equation}

\subsubsection{Rotational Dynamics}

The attitude dynamics of the quadrotor on $SO(3)$ are described by:
\begin{align}
\dot{R} &= R \omega^{\times} \label{eq:attitude_kinematics}\\
J\dot{\omega} &= \tau - \omega^{\times}(J\omega) \label{eq:attitude_dynamics}
\end{align}
where $\omega \in \mathbb{R}^3$ is the angular velocity expressed in $\mathcal{F}_B$, 
$J \in \mathbb{R}^{3 \times 3}$ is the moment of inertia matrix, 
$\tau \in \mathbb{R}^3$ represents the control torque vector, 
and $\omega^{\times} \in \mathfrak{so}(3)$ denotes the skew-symmetric operator such that $\omega^{\times}b = \omega \times b$ for any $b \in \mathbb{R}^3$.

\begin{remark}
Since the UAV attitude is fully actuated by three independent torques, an inner-loop controller can be assumed to ensure $R \to R_d$~\cite{10149091}. The commanded attitude, expressed by the roll--pitch--yaw angles $(\phi_d,\theta_d,\psi_d)$, is related to the control input $u = FRe_3$, with $F=\|u\|$. Specifically as given \cite{kendoul2009nonlinear}, $\phi_d = \arcsin\!\big((u_x \sin\psi_d - u_y \cos\psi_d)/F\big)$ and $\theta_d = \arctan\!\big((u_x \cos\psi_d + u_y \sin\psi_d)/u_z\big)$, while $\psi_d$ is freely assigned. Assuming $u$ evolves continuously, these angles remain trackable. Hence, our focus is on the outer-loop design for stabilizing the UAV--payload dynamics via $u$, while the inner loop guarantees the required attitude tracking.
\end{remark}

The following properties and assumptions are essential for the subsequent control design:

\begin{assumption}
The payload swing angles are bounded: $(\alpha,\beta)\in(-\pi/2,\pi/2)$, ensuring that the load remains below the quadrotor and preventing cable entanglement. This is a reasonable and widely adopted assumption in the UAV–payload literature~\cite{8038248,9456981, 8825990}.
\label{assum:swing_bound}
\end{assumption}

\begin{property}
The system exhibits a cascaded underactuated structure where the translational subsystem has five DOFs with three virtual control inputs, while the rotational subsystem is fully actuated with three torque inputs.
\label{prop:cascade}
\end{property}

The primary control objective is to design control inputs $(F, \tau)$ that keep the system within the safe set while asymptotically stabilizing the quadrotor and payload system to a desired configuration and suppressing payload oscillations:
\begin{equation}
\lim_{t \to \infty} [\xi(t), \gamma(t), \dot{\xi}(t), \dot{\gamma}(t)] = [\xi_d, 0, 0, 0],
\label{eq:control_objective}
\end{equation}
\begin{equation}
[\xi_d, 0, 0, 0] \in \mathrm{int}(\mathcal{C}),
\label{eq:desired_safe}
\end{equation}
where $\xi_d \in \mathbb{R}^3$ is the desired quadrotor position and $\mathcal{C} := \{\, \mathbf{x} \mid h_{i,j}(\mathbf{x},t) \ge 0,\ \forall i,j,\ u \in \mathcal{U} \,\}$ denotes the safe set.

\begin{figure}[t]
    \centering
    \includegraphics[width=1\linewidth]{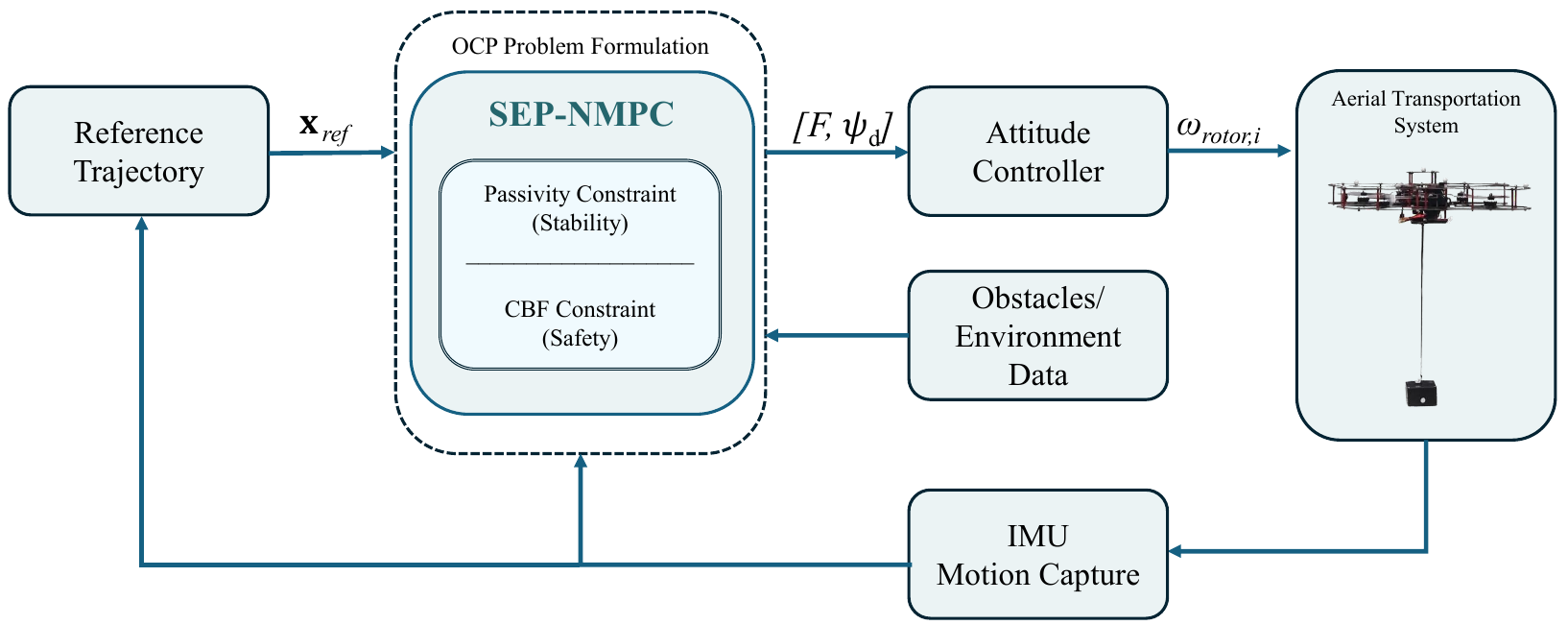}
    \vspace{-20pt}
   \caption{Block diagram of the proposed SEP-NMPC framework, where state and obstacle data are fed into the optimal control problem (OCP) enforcing passivity (stability) and HOCBF (safety) constraints to generate control inputs for safe, stable trajectory tracking.}
    \label{fig:sep_nmpc_block}
\vspace{-10pt}
\end{figure}

\section{Safety Enhanced Passivity-Based NMPC (SEP-NMPC)}

This section details the design of the proposed SEP-NMPC framework, which simultaneously enforces system stability and constraint satisfaction through energy-aware damping and control barrier functions (CBFs), respectively. The formulation is tailored for the UAV–payload dynamics described in Section~II. The overall control architecture is illustrated in Fig.~\ref{fig:sep_nmpc_block}, where state measurements and reference commands are processed through the passivity-based shaping module and the HOCBF-based safety filter before entering the NMPC optimizer.

\subsection{SEP-NMPC Formulation}
\label{subsec:sep_nmpc_formulation}

At each sampling instant $t_k$, the proposed SEP-NMPC solves:
\begin{subequations}
\label{eq:sep_nmpc}
\begin{align}
\min_{\mathbf{x}(\cdot),\,u_a(\cdot)}\;
& \begin{aligned}[t]
J \;=\; \int_{t_k}^{t_k+T} \!\Big(\, \|\mathbf{x}(t)-\mathbf{x}_d(t)\|_{Q}^{2} + \|u_a(t)\|_{R}^{2} \Big)\, dt 
\end{aligned}
\label{eq:sep_cost}
\\[0pt]
\text{s.t. } \;
& \dot{\mathbf{x}}(t) \;=\; f\big(\mathbf{x}(t),u_a(t)\big), 
\qquad \mathbf{x}(t_k)=\mathbf{x}_k,
\label{eq:sep_dynamics}
\\[2pt]
&(\alpha(t),\beta(t))\in\mathcal{A}, \quad u_a(t)\in\mathcal{U},
\\[2pt]
& u_a(t)^{\!\top} v(t) \;\le\; -\,\rho\,\|v(t)\|^{2} \;-\; \varepsilon\,\|u_a(t)\|^{2},
\label{eq:sep_passivity}
\\[2pt]
& A_{\mathrm{CBF}}\!\big(\mathbf{x}(t),t\big)\,u_a(t) \;\ge\; b_{\mathrm{CBF}}\!\big(\mathbf{x}(t),t\big).
\label{eq:sep_cbf_affine}
\end{align}
\end{subequations}
Here, $\mathbf{x}(t)\in\mathbb{R}^n$ denotes the full system state, including quadrotor position, attitude, payload swing angles, and their velocities. 
The optimization variable $u_a(t)$ is the shaped control input introduced in Section~\ref{subsec:passivity}, where its construction is detailed. 
The set $\mathcal{A}$ specifies admissible swing-angle limits $(\alpha,\beta)$, while $\mathcal{U}$ encodes actuator saturation bounds. 
The collocated output is defined as $v(t)=\dot{\xi}(t)$, and the inequality~\eqref{eq:sep_passivity} enforces a strict passivity condition with user-chosen gains $\rho,\varepsilon>0$, ensuring system energy dissipation. 
Finally, the affine inequalities~\eqref{eq:sep_cbf_affine} encode the stacked
high-order CBF constraints $h_{i,j}(\mathbf{x},t)$ for all obstacle–body pairs. The
matrices $A_{\mathrm{CBF}}$ and $b_{\mathrm{CBF}}$ are constructed by stacking
these conditions for each obstacle $i$ and body $j\in\{Q,L\}$; in our setup,
$i=1$ corresponds to the static obstacle with position $p_1(t)$, and $i=2$ to
the dynamic obstacle with position $p_2(t)$ and velocity $v_2(t)$. Since the
resulting inequalities are affine in the control input $u_a$, they fit directly
into the QP structure of the NMPC solver, and the non-negativity of each HOCBF value
ensures forward invariance of the corresponding safe sets.

\subsection{Passivity-Based Stability Constraint}
\label{subsec:passivity}

Passivity-based NMPC was first formalised by Raff \emph{et al.}~\cite{Raff2007}, who showed that for passive systems with a continuously differentiable storage function, local asymptotic stability can be achieved by embedding a passivity inequality in the optimization. We follow this paradigm, adopting the underactuated energy-shaping construction in \cite{9854092}, enforcing the passivity constraint along the prediction horizon and constructing a storage function that captures both the payload’s swing energy and convergence to the desired position.

Let $\xi_d(t)\in\mathbb{R}^3$ denote the desired position and define the translational error $e_\zeta := \xi - \xi_d$, where $K\in\mathbb{R}^{3\times 3}$ is symmetric positive definite. 
Define the state vector $\mathbf{x} := [q^\top,\dot q^\top]^\top$. 
The enhanced storage function is
\begin{equation}
V(\mathbf{x}) =
\frac{1}{2} \dot{q}^\top M(q) \dot{q}
+ m_L g l \big(1 - \cos \alpha \cos \beta \big)
+ \frac{1}{2} e_\zeta^\top K e_\zeta .
\label{eq:storage}
\end{equation}

The additional term involving $e_\zeta$ shapes the energy landscape to promote convergence to the reference position.

Using the system dynamics in \eqref{eq:trans_dynamics} and the standard identity
\begin{equation}
\dot{q}^T\bigl(\tfrac12\dot M(q)-C(q,\dot q)\bigr)\dot q=0
\end{equation}

The shaped control input is given by $u_a = u + \operatorname{col}(K e_\zeta,\,0,\,0)$, which yields the port relation $\dot{V}(\mathbf{x}) = v^\top u_a$, where the collocated output is $v(t) = \dot{\xi}(t)$.

To ensure energy dissipation, the NMPC includes the strict passivity constraint
\begin{equation}
u_a^T(t) v(t) \leq -\rho \|v(t)\|^2 - \varepsilon \|u_a(t)\|^2 ,
\quad \rho,\varepsilon > 0,
\label{eq:passivity_final}
\end{equation}
which implies $\dot V(\mathbf{x})\le 0$, so the storage function is nonincreasing and bounded from below. 
Since $\rho,\varepsilon>0$, the equality $\dot V(\mathbf{x})=0$ can hold only if
$v=\ 0$ and $u_a=\ 0$.
Following the strict passivity argument in~\cite{9854092}, this condition ensures that the largest invariant subset of $\{\mathbf{x}:\dot V(\mathbf{x})=0\}$ corresponds to the equilibrium characterized in Section~\ref{subsec:stability}. 
Therefore, by LaSalle’s invariance principle and the strict passivity condition, all closed-loop trajectories converge asymptotically to the desired equilibrium, ensuring convergence to the reference while maintaining bounded payload motion.

\subsection{Closed-Loop Stability Analysis}
\label{subsec:stability}

Consider the storage function $V(\mathbf{x})$ in \eqref{eq:storage}. From Section~\ref{subsec:passivity}
we have the port relation $\dot V(\mathbf{x})= v^\top u_a$ with the collocated output $v=\dot\xi$.
Together with the strict passivity inequality \eqref{eq:passivity_final}, it follows that
\begin{equation}
\dot{V}(\mathbf{x}) \leq -\rho \|v\|^2 - \varepsilon \|u_a\|^2 \leq 0,
\qquad \rho, \varepsilon > 0.
\end{equation}
Hence $V$ is nonincreasing along closed-loop trajectories.

\paragraph*{Boundedness via sublevel sets}
On the angle-bounded domain of Assumption~\ref{assum:swing_bound}, $V$ is positive definite about hover and radially unbounded in $(e_\zeta,\dot q)$. Every sublevel set $\Omega_c=\{\,\mathbf{x}\mid V(\mathbf{x})\le c\,\}$ is compact and forward invariant, so closed-loop trajectories are precompact.

\paragraph*{Inner-loop tracking}
We adopt the standard inner-loop assumption used in passivity-based MPC. The attitude–thrust loop renders $(F_d,R_d)\!\mapsto\!F$ exponentially accurate, or equivalently input–output passive with indices that, combined with those in \eqref{eq:passivity_final}, yield a strictly passive interconnection. At equilibrium $R\mathbf e_3$ aligns with $\mathbf e_3$ and $F=(m_Q+m_L)g\,\mathbf e_3$.

\paragraph*{Invariant set characterization}
Let $\mathcal E=\{\,\mathbf{x}\in\Omega_c\mid \dot V(\mathbf{x})=0\,\}$. From \eqref{eq:passivity_final}, $\dot V(\mathbf{x})=0$ implies
\[
v=\dot\xi=\mathbf 0 \quad\text{and}\quad u_a=\mathbf 0.
\]

To remain in $\mathcal E$, these equalities must hold for all $t\ge 0$.
Hence, on the largest invariant subset $\mathcal M\subseteq\mathcal E$, we also have $\ddot\xi=\mathbf 0$.

Using \eqref{eq:trans_dynamics} with $u=\big[F^\top,0,0\big]^\top$ and $\gamma=[\alpha,\beta]^\top$, the translation equalities are
\begin{subequations}
\label{eq:trans_three_rows}
\begin{align}
& m_L l\!\left(\cos\alpha\cos\beta\,\ddot\alpha - \sin\alpha\sin\beta\,\ddot\beta\right)
\;+\; c_{14}\dot\alpha + c_{15}\dot\beta \;=\; 0, \label{eq:Xrow}\\
& m_L l\!\left(\cos\beta\,\ddot\beta\right) \;+\; c_{25}\dot\beta \;=\; 0, \label{eq:Yrow}\\
& m_L l\!\left(\sin\alpha\cos\beta\,\ddot\alpha + \cos\alpha\sin\beta\,\ddot\beta\right)
\;+\; c_{34}\dot\alpha + c_{35}\dot\beta \;=\; 0. \label{eq:Zrow}
\end{align}
\end{subequations}

On $\mathcal M$ we have $v\equiv 0$ and $\ddot\xi\equiv 0$, and the vertical thrust balance $F=(m_Q{+}m_L)g\,\mathbf e_3$ holds, so \eqref{eq:trans_three_rows} is satisfied for all $t$.

From \eqref{eq:Yrow} and the explicit $c_{25}$,
\begin{equation}
\cos\beta\,\ddot\beta - \dot\beta^{2}\sin\beta = 0.
\label{eq:Y_simplified}
\end{equation}

Eliminating $\ddot\beta$ from \eqref{eq:Xrow}–\eqref{eq:Zrow} by the combination
$\cos\alpha\,\eqref{eq:Xrow} + \sin\alpha\,\eqref{eq:Zrow}$ gives
\begin{equation}
\cos\alpha\,\ddot\alpha - \dot\alpha^{2}\sin\alpha = 0.
\label{eq:alpha_simplified}
\end{equation}

On the domain $|\alpha|,|\beta|<\tfrac{\pi}{2}$ we have $\cos\alpha>0$ and $\cos\beta>0$.
Equations \eqref{eq:Y_simplified}–\eqref{eq:alpha_simplified} can be rewritten as
\[
\frac{d}{dt}(\dot\beta\cos\beta)=0,
\qquad
\frac{d}{dt}(\dot\alpha\cos\alpha)=0.
\]

Hence $\dot\beta\cos\beta=c_1$ and $\dot\alpha\cos\alpha=c_2$ for constants $c_1,c_2$.
Since closed-loop trajectories remain in the compact invariant set $\Omega_c$,
the swing angles must remain bounded for all time. This is possible only if
$c_1=c_2=0$, as otherwise $\beta(t)$ or $\alpha(t)$ would grow unbounded.
Therefore, $\dot\alpha=\dot\beta=0$ on the largest invariant set $\mathcal M\subset\Omega_c$,
following the same invariant-set reasoning used for UAV slung-payload systems in \cite{8038248}.

With $\dot\alpha=\dot\beta=0$, the last two equations reduce to
\begin{subequations}
\label{eq:pend_rest}
\begin{align}
m_L l^2 \cos^2 \beta \, \ddot{\alpha} &= - m_L g l \, \sin \alpha \, \cos \beta, \\
m_L l^2 \, \ddot{\beta} &= - m_L g l \, \cos \alpha \, \sin \beta .
\end{align}
\end{subequations}

Since $\cos\alpha>0$ and $\cos\beta>0$ on the domain, \eqref{eq:pend_rest} forces $\sin\alpha=\sin\beta=0$.
Thus $(\alpha,\beta)=(0,0)$ and $(\ddot\alpha,\ddot\beta)=(0,0)$ on $\mathcal M$.

By the inner-loop assumption, the equilibrium thrust equals $(m_Q+m_L)g\,\mathbf e_3$, ensuring $e_\zeta=\mathbf 0$. Together with the invariant-set conditions above, the largest invariant subset of $\mathcal E$ is $\mathcal M=\{(\xi,\gamma,\dot\xi,\dot\gamma)=(\xi_d,0,0,0)\}$. By LaSalle’s invariance principle, all trajectories converge asymptotically to this equilibrium, i.e., $\xi(t)\to\xi_d$ and $(\dot\xi,\alpha,\beta)\to 0$.

\newcommand{\Aij}{A^{(i,j)}_{\mathrm{CBF}}}
\newcommand{\bij}{b^{(i,j)}_{\mathrm{CBF}}}

\subsection{Safety via High-Order Control Barrier Functions}
\label{subsec:cbf}

To ensure collision-free navigation in cluttered environments, we employ high-order control barrier functions (HOCBFs)~\cite{7782377,9516971} that explicitly account for both the quadrotor and the suspended payload. Unlike standard CBF formulations that consider only the vehicle body, our approach recognizes that the swinging payload substantially enlarges the system’s effective footprint, particularly during aggressive maneuvers. All barrier inequalities are written directly in terms of \(u_a\), keeping them affine and QP-compatible.

\paragraph*{Control-affine translational channel}
Let \(p_Q=\xi\) denote the quadrotor position and \(v=\dot\xi\) its velocity. The translational dynamics of the outer loop are expressed in control-affine form as
\begin{equation}
\dot p_Q = v,\qquad \dot v = f_v(\mathbf{x}) + G_v(\mathbf{x})\,u_a,
\label{eq:trans_affine}
\end{equation}
where \(f_v(\mathbf{x})\) denotes the passive dynamics (i.e., the natural motion due to gravity, coupling, and swing effects in the absence of input) and \(G_v(\mathbf{x})\in\mathbb{R}^{3\times 3}\) is the input map from the shaped control input \(u_a\) to acceleration. For the payload, the acceleration can be written as \(\ddot p_L=\ddot p_Q+\eta_L(\mathbf{x})\), where \(\eta_L(\mathbf{x})\) collects the additional swing-induced terms; crucially, the Jacobian $\partial \ddot p_L / \partial u_a$ is the same \(G_v(\mathbf{x})\). This ensures that both the quadrotor and payload channels are affine in the decision variable \(u_a\).

\paragraph*{Dual-body clearance and effective radii}
Define the body positions as
\[
p_Q := \xi, \qquad
p_L := \xi + l\,[\sin\alpha\cos\beta,\; \sin\beta,\; -\cos\alpha\cos\beta]^\top,
\]
where $l$ is the cable length. For obstacle $i$ with center $p_{o,i}(t)$ and radius $R_i$, choose inflated body radii $r_j$ for $j\in\{Q,L\}$ and a safety margin $\Delta>0$, and set
\[
d_{\min,i,j} := R_i + r_j + \Delta.
\]

Let $r_{i,j}(t) := p_j(t) - p_{o,i}(t)$ for $j\in\{Q,L\}$, so the squared-clearance functions are

\[
h_{i,j}(\mathbf{x},t) = \|r_{i,j}(t)\|^2 - d_{\min,i,j}^2,
\label{eq:cbf_clearance}
\]
with safe sets $\mathcal{C}_{i,j}=\{\,\mathbf{x} \mid h_{i,j}(\mathbf{x},t)\ge 0\,\}$.

\paragraph*{Dynamic obstacles}
When obstacle \(i\) moves with velocity \(v_{o,i}(t)\) and acceleration \(\dot v_{o,i}(t)\), the derivatives of the squared-clearance functions are
\begin{align}
\dot h_{i,j}(\mathbf{x},t) &= 2\,r_{i,j}(t)^{\!\top}\big(\dot p_j(t)-v_{o,i}(t)\big),
\label{eq:h_dot}\\
\ddot h_{i,j}(\mathbf{x},t) &= 2\|\dot p_j(t)-v_{o,i}(t)\|^2
+ 2\,r_{i,j}(t)^{\!\top}\big(\ddot p_j(t)-\dot v_{o,i}(t)\big).
\label{eq:h_ddot}
\end{align}

Thus, obstacle motion enters explicitly as a time-varying exogenous signal~\cite{9483029}, while the control input \(u_a\) influences only through the acceleration channel.

\paragraph*{Relative degree two and HOCBF recursion}
A key observation is that the functions \(h_{i,j}(\mathbf{x},t)\) depend only on position, while the input \(u_a\) acts through acceleration. This induces a relative degree of two with respect to \(u_a\), making standard first-order CBFs insufficient~\cite{9516971,9456981}. To address this, we construct a second-order CBF recursion with positive gains $\kappa_1,\kappa_2>0$:
\begin{subequations}
\label{eq:cbf_recursion}
\begin{align}
\psi_{i,j,1}(\mathbf{x},t) &= \dot h_{i,j}(\mathbf{x},t)+\kappa_{1}\,h_{i,j}(\mathbf{x},t), \label{eq:cbf_psi1}\\
\psi_{i,j,2}(\mathbf{x},u_a,t) &= \dot\psi_{i,j,1}(\mathbf{x},t)+\kappa_{2}\,\psi_{i,j,1}(\mathbf{x},t)\;\ge 0. \label{eq:cbf_psi2}
\end{align}
\end{subequations}
Substituting
\[
\ddot p_j=f_{v,j}(\mathbf{x},t)+G_v(\mathbf{x})\,u_a ,
\]
where \(f_{v,j}(\mathbf{x},t)\) collects the passive dynamics of body \(j\), yields an affine inequality in the control input.

\paragraph*{QP form and affine representation}
Expanding \eqref{eq:cbf_psi2}, we obtain
\begin{equation}
\label{eq:psi2_affine}
\begin{aligned}
\psi_{i,j,2}(\mathbf{x},u_a,t)
&= L_f^{2}h_{i,j}(\mathbf{x},t) + \kappa_1 L_f h_{i,j}(\mathbf{x},t) 
   + \kappa_2 \psi_{i,j,1}(\mathbf{x},t) \\
&\quad + 2\,r_{i,j}(t)^{\!\top}G_v(\mathbf{x})\,u_a \;\ge 0
\end{aligned}
\end{equation}

The corresponding affine constraint can thus be written per pair as $A^{(i,j)}_{\text{CBF}}(\mathbf{x},t)\,u_a \ge b^{(i,j)}_{\text{CBF}}(\mathbf{x},t)$, where
\begin{align*}
\Aij(\mathbf{x},t) &:= 2\,r_{i,j}(t)^{\!\top}G_v(\mathbf{x}), \\[3pt]
\bij(\mathbf{x},t) &:= -\Big(L_f^{2} h_{i,j}(\mathbf{x},t)
+\kappa_{1} L_f h_{i,j}(\mathbf{x},t) \\[-1pt]
&\hphantom{:=}\quad
+\kappa_{2}\psi_{i,j,1}(\mathbf{x},t)\Big).
\end{align*}

\begin{figure}[t]
  \centering
  \includegraphics[width=1.01\linewidth]{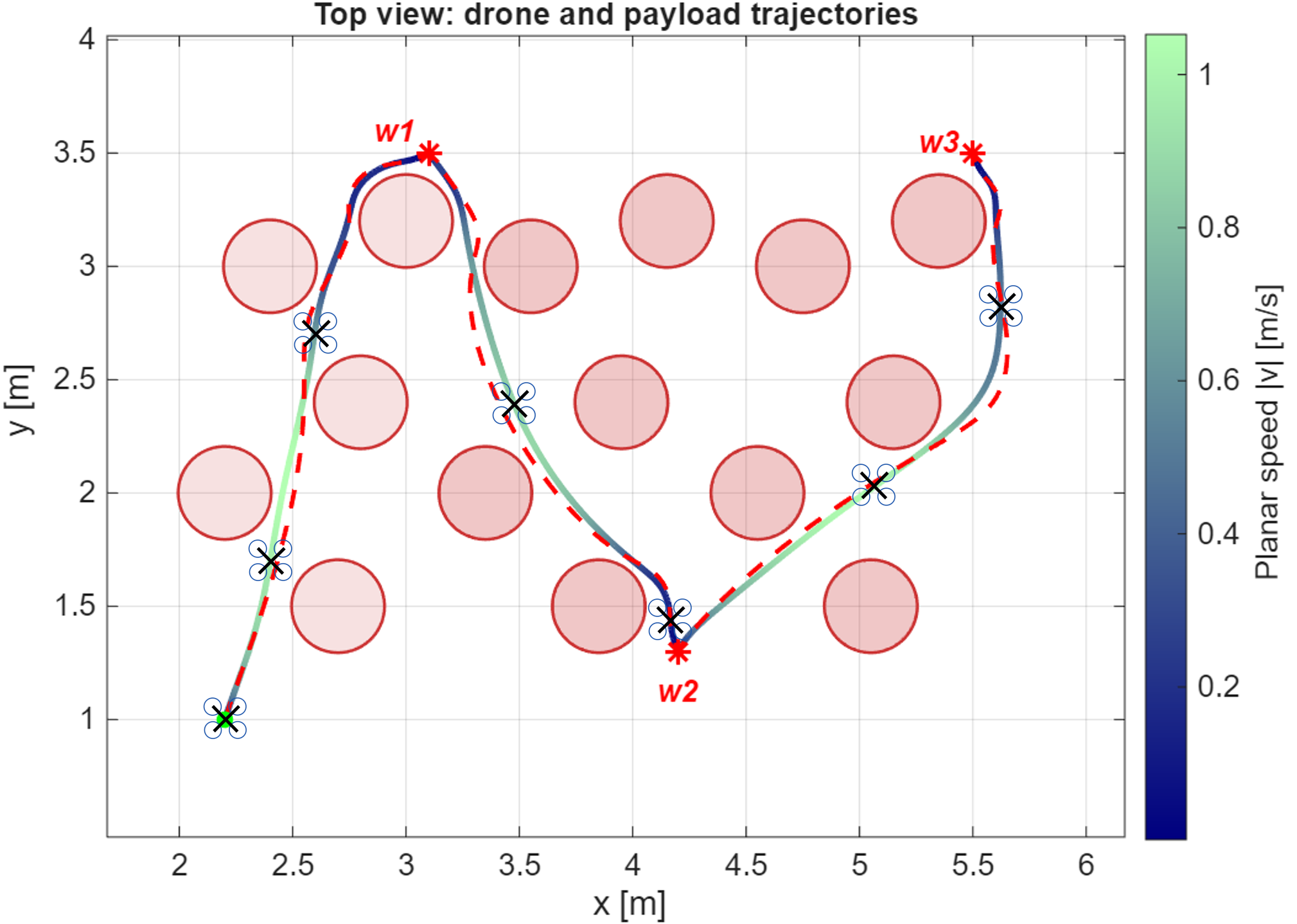}
\caption{Top-view trajectories of the quadrotor–payload system. The quadrotor path is a solid curve color-coded by planar speed (see colorbar), while the payload path is the red dashed curve. Circular red disks denote obstacle positions; waypoints \(w_1,w_2,w_3\) are marked.}

  \label{fig:static_gate}
\vspace{-7pt}  
\end{figure}

\begin{figure*}[t]
  \centering
  \begin{minipage}{0.49\linewidth}
    \centering
    \includegraphics[width=0.82\linewidth]{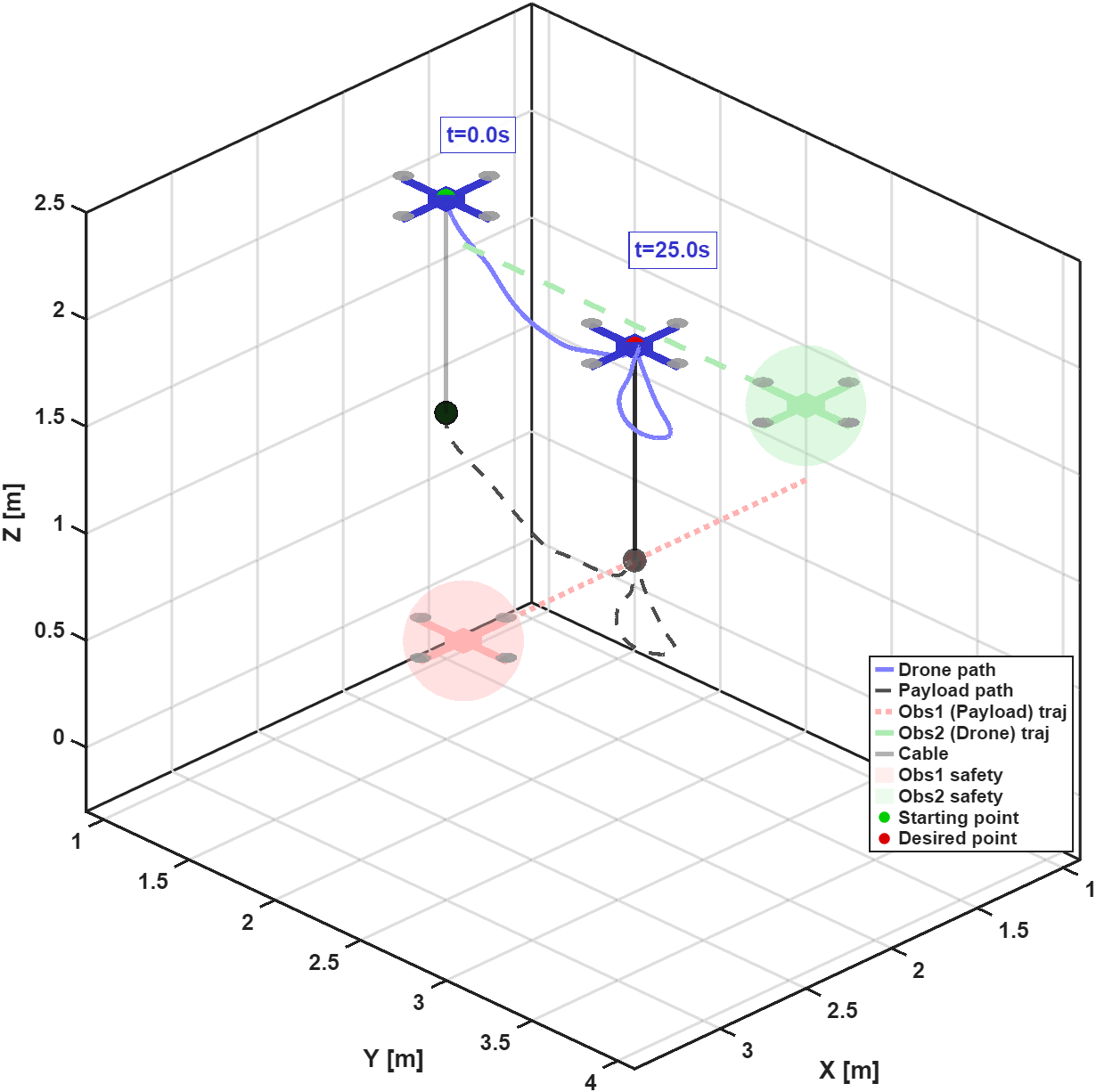}
  \end{minipage}\hfill
  \begin{minipage}{0.49\linewidth}
    \centering
    \includegraphics[width=\linewidth]{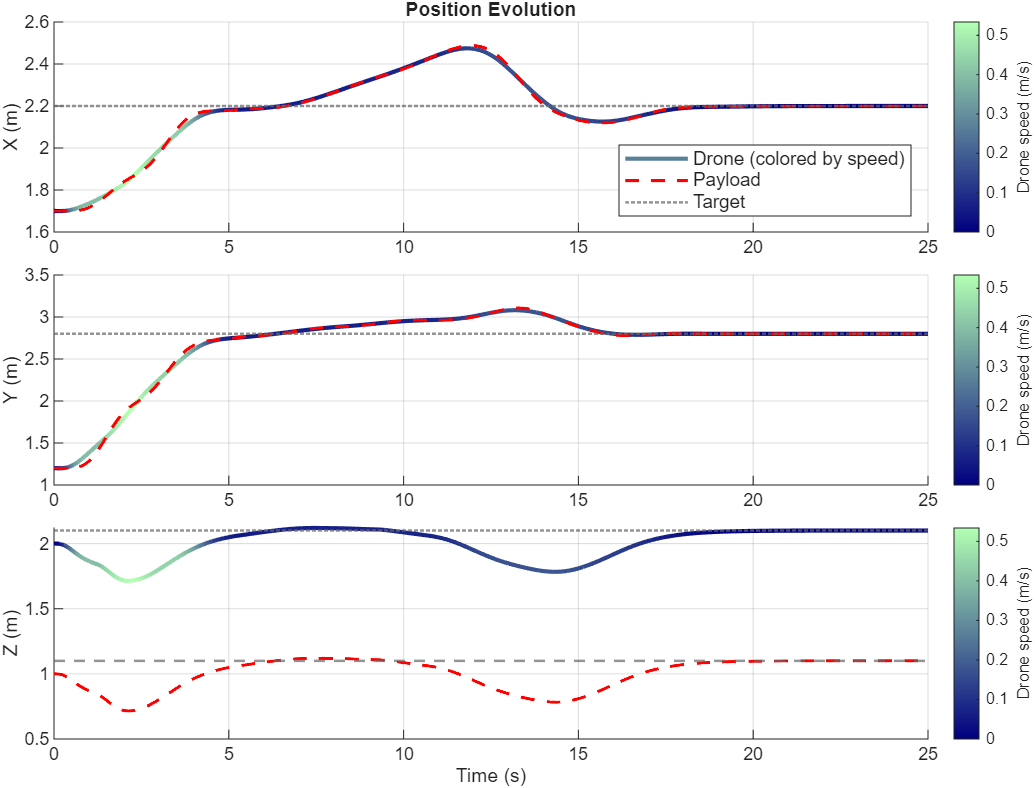}
  \end{minipage}
  \caption{Dynamic obstacle scenario. 
  Left: Reciprocal avoidance between the ego quadrotor–payload system (black/grey) and two moving intruder drones (red/green), with dashed lines indicating the maintained safety corridors. 
  Right: Position trajectories of the quadrotor (blue), payload (red), and target (green, dotted) along the $x$-, $y$-, and $z$-axes.}
  \label{fig:dyn_obs_pos}
  \vspace{-10pt}
\end{figure*}

At each sequential quadratic programming (SQP) linearization point (i.e., we linearize the dynamics and constraints and solve a QP subproblem), the inequalities \eqref{eq:psi2_affine} for all obstacle–body pairs are stacked into
\begin{equation}
A_{\mathrm{CBF}}(\mathbf{x},t)\,u_a \;\ge\; b_{\mathrm{CBF}}(\mathbf{x},t),
\label{eq:cbf_stack}
\end{equation}
where each row corresponds to one pair \((i,j)\). This preserves linearity in \(u_a\) and the QP structure of the optimization step.

\paragraph*{Forward invariance guarantee}
Suppose Assumption~\ref{assum:swing_bound} holds and the dynamics \eqref{eq:trans_affine} are locally Lipschitz. If \(h_{i,j}(\mathbf{x}_0,t_0)\ge 0\) for all obstacles \(i\) and bodies \(j\in\{Q,L\}\), then any input \(u_a(\cdot)\) satisfying \eqref{eq:cbf_stack} renders the composite safe set \(\bigcap_{i,j}\mathcal{C}_{i,j}\) forward invariant. In other words, both the quadrotor and payload maintain clearance from all obstacles for all future times.

The passivity indices \(\rho,\varepsilon\) and the HOCBF gains \(\kappa_1,\kappa_2\) balance aggressiveness and conservatism. Larger \(\rho\) and \(\varepsilon\) increase damping and limit control effort, yielding smoother but more conservative responses, while larger \(\kappa_1,\kappa_2\) trigger earlier avoidance and increase clearance margins at the cost of longer paths. In all experiments, parameters were selected to ensure strict feasibility and maintain real-time execution at 100\,Hz.

\section{Simulation and Experimental Validation}%
\label{sec:simulation}

We now validate the proposed SEP-NMPC framework through a comprehensive set of simulations and real-world experiments. The objective of this section is twofold: (i) to demonstrate that the controller enforces both passivity-based stability and high-order CBF safety simultaneously, and (ii) to assess its real-time feasibility under scenarios representative of cluttered UAV operations. To this end, we designed two complementary case studies. First, a static-obstacle avoidance task highlights the ability of SEP-NMPC to regulate payload swing while preserving clearance margins. Second, a dynamic-obstacle interaction introduces moving intruders governed by the same law, thereby testing reciprocal safety in multi-agent settings. 

Across both studies, we quantify safety through minimum separation distances, stability through payload swing suppression and trajectory tracking accuracy, and efficiency through average solver times. These metrics allow direct comparison against baseline formulations such as first-order CBFs, which enforce safety using a velocity-level barrier condition on the clearance function, as well as state constraints and passivity-only controllers. In addition to simulation, the same scenarios were implemented on a quadrotor platform carrying a cable-suspended payload, thereby bridging theory with real-world flight tests. The combined results verify that SEP-NMPC consistently achieves safe obstacle avoidance and stable trajectory tracking in real time, outperforming baseline strategies.

\begin{figure}[!t]
  \centering
  \includegraphics[width=0.95\linewidth]{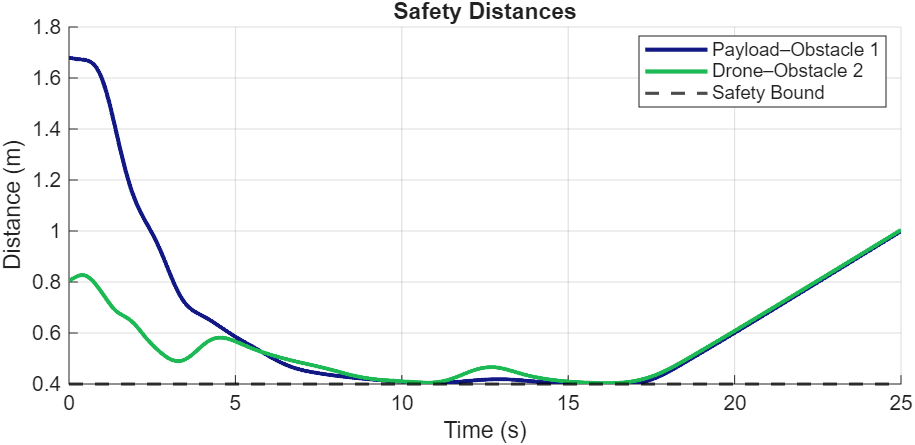}
  \vspace{-7pt}  
  \caption{Representative safety margins during dynamic obstacle avoidance: (a) payload-obstacle~1 and (b) drone-obstacle~2, both compared against the enforced minimum distance $d_{\min}$.}
  \label{fig:safety_distances}
\vspace{-30pt}  
\end{figure}

\subsection{Software Setup}%
All simulations were executed in \textsc{MATLAB} R2025a on Windows.  
The optimal-control problem was formulated and solved online with \texttt{acados}~\cite{Verschueren2021} via the \texttt{CasADi}~\cite{Andersson2019} interface, using a fixed-step fourth-order Runge–Kutta integrator.  
A prediction horizon of $T = 2\,$s was discretised into $N = 40$ shooting nodes, and solved using the SQP-RTI solver with warm starts, which evaluates a single sequential quadratic programme each control cycle.  
All runs were performed on a laptop equipped with an Intel\textregistered{} Core\texttrademark{} i7-11800H processor.

\begin{figure*}[t]
  \centering
      \vspace{+8pt}

  \includegraphics[width=0.96\linewidth]{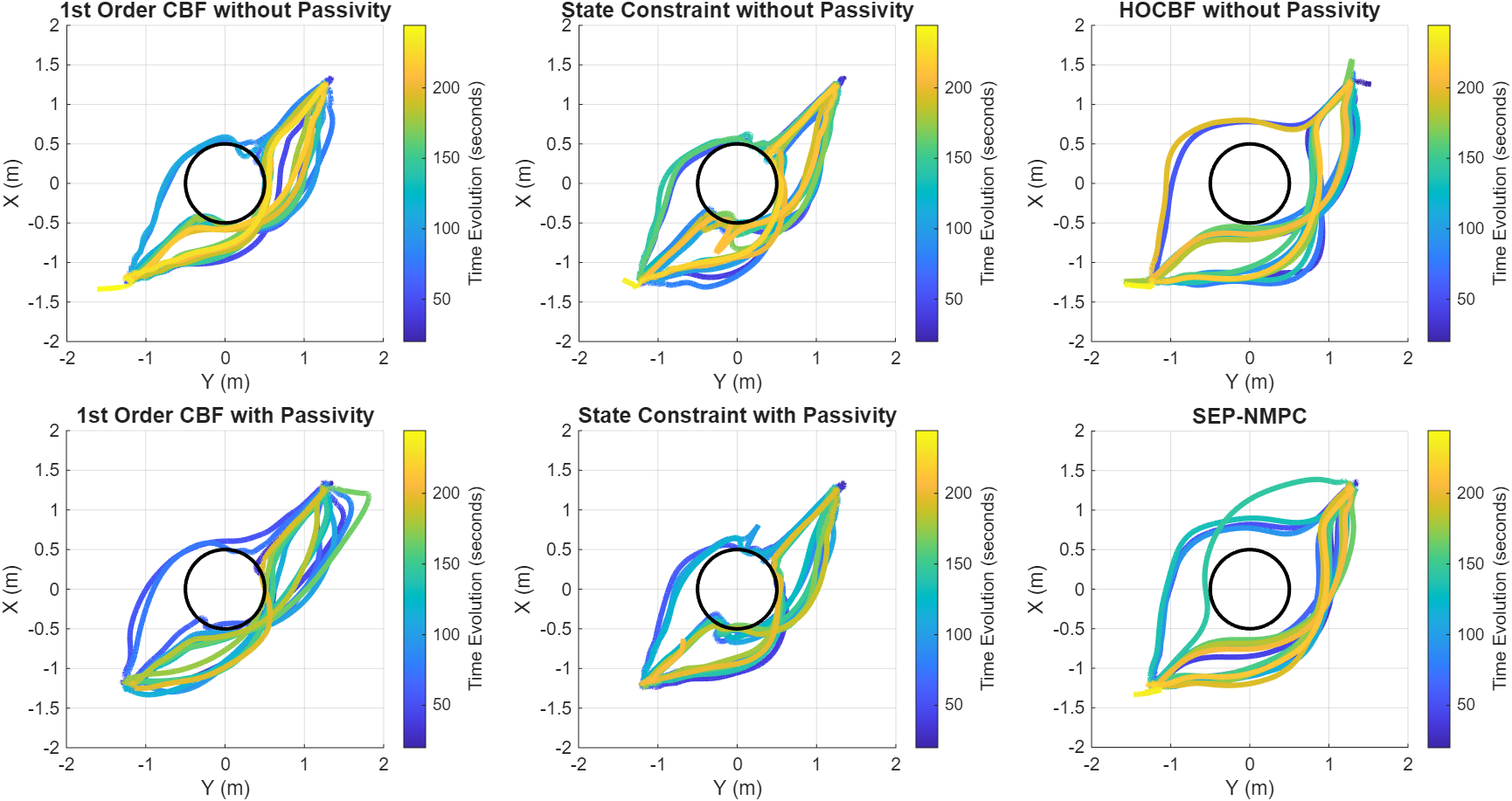}
    \vspace{-8pt}
\caption{Top-view trajectories from real-world experiments with a single obstacle. 
Each subplot shows 20 repeated trials, color-coded by time. 
Top row: (left) 1st-order CBF without passivity, (middle) state-constraint without passivity, (right) high-order CBF (HOCBF) without passivity. 
Bottom row: (left) 1st-order CBF with passivity, (middle) state-constraint with passivity, (right) proposed SEP--NMPC with high-order CBF and passivity. 
Without passivity, both CBF and state-constraint approaches show boundary violations and inconsistent convergence. 
HOCBF restores feasibility but overshoot persists. 
Adding passivity improves smoothness and safety, while SEP-NMPC achieves the most reliable avoidance and convergence.}
  \label{fig:exp_topviews}
  \vspace{-7pt}

\end{figure*}

\subsection{Static and Dynamic Obstacle Scenarios}%
To assess simultaneous safety and stability, the quadrotor–payload system was commanded to navigate between static cylindrical obstacles while the SEP-NMPC maintained both the control barrier function safety margins and the passivity constraint. Waypoints induced appreciable payload swing before convergence, thereby stressing both components of the controller. Fig.~\ref{fig:static_gate} illustrates a representative trajectory, highlighting the maintained clearances achieved by the proposed framework.

In addition, obstacles were allowed to move rather than remain fixed. Each obstacle was modelled as a second drone; One intruder drone was commanded to cross the ego quadrotor’s desired waypoint, while a second intruder crossed the payload’s desired waypoint. Throughout the manoeuvre the control-barrier-function constraints of \emph{both} systems cooperatively enforced a safe separation, and the passivity constraint ensured stable motion while reacting to nearby drones.
Fig.~\ref{fig:dyn_obs_pos} shows a representative interaction, in which the ego quadrotor smoothly adjusts its path while the intruders execute complementary avoidance trajectories. Average solver time in simulation remained below 10\,ms, confirming real-time feasibility even under dynamic interactions. In addition, {Fig.~\ref{fig:safety_distances} illustrates the corresponding safety margins, verifying that both the payload and drone maintained clearance above the prescribed bound $d_{\min}$.

\begin{table}[t]
\centering
\renewcommand{\arraystretch}{1.1}
\caption{Ablation on single-obstacle case (20 runs).}
\label{tab:ablation}

\scriptsize
\begin{tabular}{l c c c}
\toprule
\toprule
Method & Viol./Infeas. & Overshoot (\#) & Solve [ms] \\
\midrule
State Constraint only        & 16 / 19 & 0 & 5.26 \\
State Constraint + Passivity & 14 / 17 & 0 & 7.75 \\
1st-Order CBF only           & 6 / 18  & 0 & 4.90 \\
1st-Order CBF + Passivity    & 6 / 8   & 3 & 7.52 \\
High-Order CBF only          & \textbf{0 / 0} & 4 & 5.06 \\
\textbf{SEP-NMPC}            & \textbf{0 / 0} & \textbf{0} & 8.74 \\
\bottomrule
\bottomrule
\end{tabular}

\vspace{2pt}
\footnotesize \textit{Notes:}
{Viol.} counts inside $\!\to\!$ outside obstacle region cycles,
{Infeas.} is solver failure episodes, and
{Overshoot} counts goal overshoot events $>$10\,cm from the desired position.
\vspace{-10pt}
\end{table}

\subsection{Real-World Experiments}
\label{sec:experiments}

Experiments were carried out on a Quanser QDrone2 carrying a cable–suspended point-mass payload. The quadrotor has a mass of 1.5\,kg, the payload weighs 0.20\,kg, and the cable length is 0.5\,m. All computation was executed onboard using an NVIDIA Jetson Xavier NX. The proposed SEP-NMPC was implemented with \texttt{acados} and \texttt{CasADi}, running in real time at 100\,Hz with a prediction horizon of $T=2$\,s and $N=40$ shooting nodes, matching the simulation setup. A warm-started SQP-RTI scheme was used, with OpenMP parallelization across nodes to maintain low-latency updates. Obstacles were modeled as discs in the horizontal plane with an inflated safety radius of 0.5\,m, enforced simultaneously for the quadrotor and suspended payload through control barrier functions.

To benchmark performance, we compared 1st-order CBF, state-constraint, and the proposed SEP-NMPC, each tested with and without passivity, along with a passivity-free HOCBF baseline. Each configuration was flown in 20 repeated trials under the same initial conditions. Both single- and multiple-obstacle scenarios were attempted; representative single-obstacle results are reported here, while the more complex setups are included in the supplementary video.


\begin{table}[t]
\centering
\renewcommand{\arraystretch}{1.05}
\setlength{\tabcolsep}{6pt}
\caption{Experimental Results of SEP-NMPC \\ (Single-Obstacle, 20 Trials)}
\label{tab:exp_summary}
\begin{tabular}{l c}
\toprule
\toprule
Metric & Result \\
\midrule
Success rate & 20/20 \\
Min clearance (median) & 0.53 m (IQR 0.51--0.55) \\
Max swing angle ($|\alpha|,|\beta|$) & $54.94^{\circ}$ \\
Tracking RMSE$_{xyz}$ & 0.035 m (median) \\
Solver time & 8.74 ms (median) \\
Overruns $>$10 ms & 0\% \\
\bottomrule
\bottomrule
\end{tabular}
\vspace{-15pt}
\end{table}

Fig.~\ref{fig:exp_topviews} shows the time-colored top views of all six controllers in the single-obstacle case. Clear performance differences emerge: without passivity, both 1st-order CBF and state-constraint formulations exhibit frequent boundary violations and solver failures near the obstacle. HOCBFs restore feasibility but overshoot occurs when passivity is absent. Introducing the passivity inequality markedly improves safety and smoothness, reducing oscillations and ensuring trajectories remain outside the inflated safety disc. Among all cases, SEP-NMPC consistently achieves the best trade-off between safety and convergence, avoiding the obstacle while eliminating overshoot and converging smoothly to the goal. Table~\ref{tab:ablation} further quantifies these trends by reporting violation counts, infeasibility episodes, overshoot occurrences, and solver times across 20 runs. The results highlight that while state-constraint methods incur the highest number of failures, the 1st-order CBF with passivity reduces both violations and infeasibility but still suffers from occasional overshoot. This overshoot arises when the controller encounters infeasible instances: the passivity inequality prevents the optimizer from issuing overly aggressive control inputs, which in turn makes the system unable to recover quickly and induces transient instability. In contrast, HOCBFs handle feasibility more gracefully but without passivity can still overshoot the goal, underscoring the complementary nature of the two ingredients. 

Overall, these comparisons confirm that combining high-order safety constraints with passivity not only restores feasibility but also enforces a dissipative closed-loop behavior, yielding robust performance for the quadrotor slung payload system in real-world settings. Table~\ref{tab:exp_summary} further provides a detailed breakdown of the achieved performance.

\section{Conclusion}

This work introduced a Safety Enhanced Passivity-Based NMPC for quadrotor–payload transport that unifies passivity–based stability with high-order CBF safety. A shaped energy storage function and a strict passivity inequality provide a dissipative closed loop with asymptotic convergence, while HOCBFs enforce forward invariance of dual–body (vehicle and payload) clearance sets, including against moving obstacles. The resulting optimization remains QP-compatible and runs online without gain scheduling or heuristic switching, offering reliable performance even in cluttered, dynamic settings. Simulation and real-world results confirmed that SEP-NMPC achieves collision–free navigation, smooth convergence, and consistently low solver times, validating its suitability for real-time UAV transport missions. Future work will explore integrating learning-based modules for improved prediction of disturbances and obstacle motion, as well as systematic strategies to enhance controller feasibility by refining constraint handling and recovery from near-infeasible situations.

\bibliographystyle{IEEEtran}
\bibliography{references}

\end{document}